\documentclass[a4paper,twoside]{article}

\usepackage{epsfig}
\usepackage{subcaption}
\usepackage{calc}
\usepackage{amssymb}
\usepackage{amstext}
\usepackage{amsmath}
\usepackage{amsthm}
\usepackage{multicol}

\usepackage{fontspec}
\usepackage{apalike}
\usepackage{algorithm2e}
\usepackage[bottom]{footmisc}
\usepackage{SCITEPRESS}     

\usepackage{adjustbox}
\usepackage{graphicx}
\usepackage{hhline}
\usepackage[table]{xcolor}

\begin{document}

\title{Age and Gender Estimation from Face and Ocular Smartphone Images \\ via Identity Transfer and Ordinal Learning}


\author{\authorname{Fernando Alonso-Fernandez\sup{1}\orcidAuthor{0000-0002-1400-346X}, Kevin Hernandez-Diaz\sup{1}\orcidAuthor{0000-0002-9696-7843}, Jose Maria Buades\sup{2}\orcidAuthor{0000-0002-6137-9558}, Josef Bigun\sup{1}\orcidAuthor{0000-0002-4929-1262}}
\affiliation{\sup{1}School of Information Technology, Halmstad University, Sweden}
\affiliation{\sup{2}Computer Graphics and Vision and AI Group, University of Balearic Islands, Spain}
\email{\{feralo, kevin.hernandez-diaz, josef.bigun\}@hh.se, josemaria.buades@uib.es}
}


\abstract{
We investigate the transferability of identity-recognition representations to gender and age estimation from face and ocular images. 
Three CNNs of varying complexity, pre-trained for face or ocular identity recognition, are evaluated using fixed features with SVMs and end-to-end adaptation.
Experiments use 11,299 frontal smartphone images from Adience, following its predefined five-fold protocol and subject-disjoint validation.
Results show that domain-matched identity pre-training outperforms direct ImageNet and scratch initialisation in end-to-end training,
%
%
while combining both ocular regions 
%
narrows the gap with full-face estimation, especially for age.
We additionally introduce an ordinal loss that exploits the ordering of age groups and consistently improves 1-off accuracy across architectures and input regions.
Across the evaluated configurations, 
best gender accuracies reach 88.1\% for the face and 82.9\% for combined ocular regions,
while age estimation reaches 59.0\% exact and 93.1\% 1-off accuracy for ocular images, versus 57.6\% and 95.2\% for the full face.
To our knowledge, these are the highest ocular age accuracies reported on Adience.
}

\onecolumn \maketitle \normalsize \setcounter{footnote}{0} \vfill

\section{INTRODUCTION}
\label{sec:intro}

Soft-biometric attributes, such as age and gender, provide complementary information for biometric recognition, subject filtering, and attribute-dependent applications \cite{[Dantcheva16softbio],Becerra2019airFaceSoftbioForensicSurveillSurvey}.
Although most research relies on the full face, its visibility can be limited by pose, occlusion, or acquisition conditions \cite{2026PAMI_50yrfacerecognition_jain}.
The ocular or periocular region often remains visible and contains useful information for identity and soft-biometric estimation \cite{sharma23cviu_periocular_masks_survey,Alonso24computers_periSOA,2026_alonsoECCVws_periocular_softbio_soa}.
However, the reduced facial information in the ocular area makes demographic estimation more challenging.

This paper addresses age and gender estimation from frontal smartphone imagery.
We use Adience \cite{[EidingerHassner14_Adience]}, an in-the-wild dataset of Flickr images providing a more demanding setup than controlled databases \cite{2026_alonsoECCVws_periocular_softbio_soa}.
Age and gender estimation from full-face Adience images has evolved from handcrafted descriptors \cite{[EidingerHassner14_Adience],[Bhattacharyya19scGenderFacialRegionsGA]} and conventional CNNs \cite{[LeviHassner15_AgeGender],[Alonso21ietSoftBio]} to specialised architectures exploiting local/global information \cite{[Fang19neurocomputingGenderAgeMultiStage]}, attention, recurrent modules \cite{[Zhang20tcsvtFaceAgLSTM]}, and transformers \cite{2023_KuprashevichAIST_MiVOLOAgeGender,2024_SinghNCA_HybridTransformerSequencer}.
Some recent approaches also rely on large external datasets labelled for age and gender. 
In contrast, fewer works have addressed ocular soft biometrics, using CNNs
%
%
\cite{[Rattani17_AgeOcularIJCB],[Alonso21ietSoftBio],2023_GhrbanJCSSP_gender_face_eyes} and handcrafted descriptors \cite{[Bhattacharyya19scGenderFacialRegionsGA]}.
The performance gap with full-face estimation remains substantial, particularly for age. 
Direct comparisons are also complicated by differences in Adience subsets used, evaluation protocols, age-group definitions, and external training data.

Here, we investigate the transferability of biometric identity representations to age and gender estimation from full-face and ocular images.
Our contributions are threefold:
$i$) evaluating three CNNs of varying complexity pre-trained for face or ocular identity recognition, using fixed feature extraction and end-to-end adaptation, and comparing end-to-end training with direct ImageNet and scratch initialisation to assess the benefits of identity pre-training;
%
$ii$) investigating the benefits of combining both ocular regions to narrow the performance gap with full-face estimation, especially for age; and
$iii$) introducing an ordinal loss that exploits the ordering of age groups to improve age estimation.
Our approach improves previous ocular age results on the same Adience subset \cite{[Alonso21ietSoftBio]} by more than 10\% (exact accuracy) and 7-9\% (1-off accuracy), representing, to the best of our knowledge, the highest ocular age accuracies reported on Adience.

\begin{figure}[t]
\centering
\includegraphics[width=0.46\textwidth]{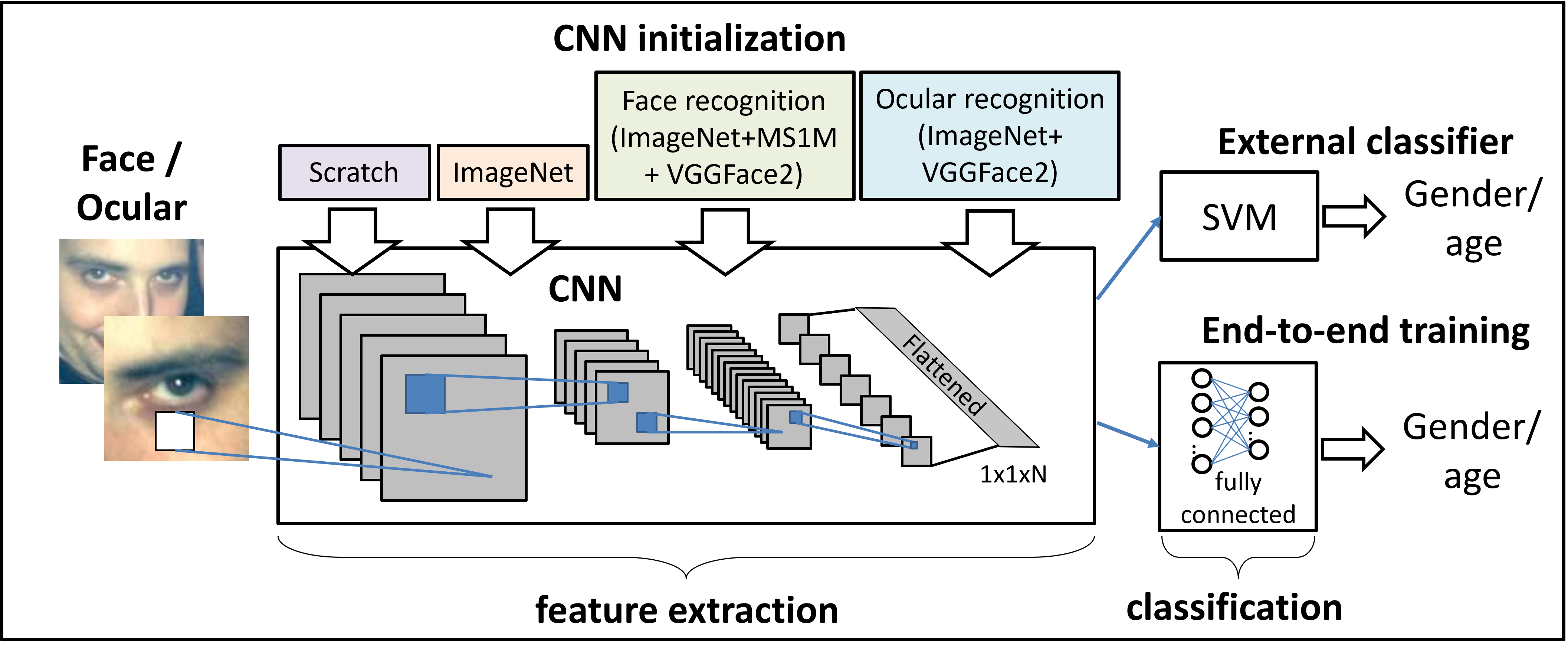}
\caption{Soft-biometrics estimation framework. Scratch and direct ImageNet initialisation are evaluated only with end-to-end training.}
\label{fig:system}
\end{figure}

\begin{figure}[t]
\centering
\includegraphics[width=0.46\textwidth]{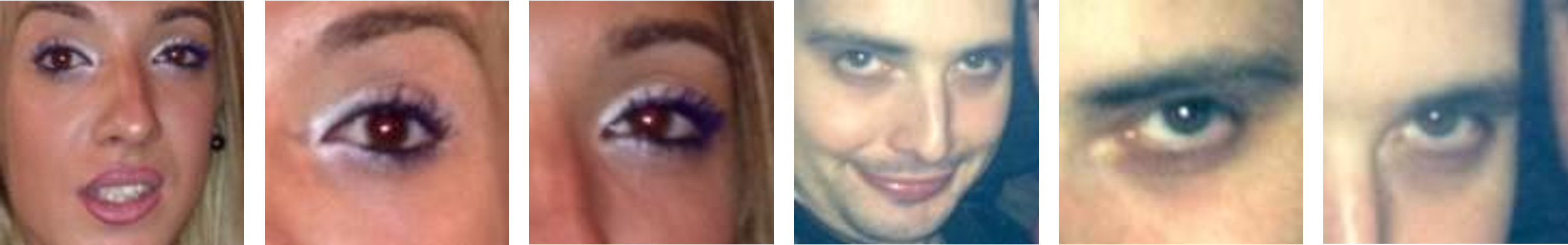}
\caption{Images from Adience \cite{[EidingerHassner14_Adience]}.}
\label{fig:adience_images}
\end{figure}

\section{NETWORKS FOR GENDER AND AGE CLASSIFICATION}
\label{sec:networks}

We evaluate three backbones of varying complexity, depth, and parameters (Table~\ref{tab:networks}): SqueezeNet \cite{[Iandola16SqueezeNet]} (light), MobileNetv2 \cite{[Sandler18mobilenetv2]} (medium) and ResNet50 \cite{[He16]} (large). 
%
%
SqueezeNet (SQ) achieves a compact architecture through squeeze-and-expand modules, MobileNetV2 (MB2) employs inverted residuals and depth-wise separable convolutions to reduce computational cost, and ResNet50 (R50) uses residual connections that bypass intermediate layers to facilitate the training of deep networks.

We use MATLAB R2024b models adapted to 113$\times$113 inputs by changing the stride of the first convolutional layer from 2 to 1.
%
%
This preserves the remaining architecture and allows reuse of ImageNet weights where applicable.
%
%
Input images are normalised by subtracting 127.5 and dividing by 128.
For SqueezeNet, we follow \cite{[Alonso20SqueezeFacePoseNet]} and add batch normalisation between convolutions and ReLU (missing in the original model).

\begin{table}[]
\centering

\caption{Networks evaluated in this paper. The vector size corresponds to the layer preceding the classification layer in each CNN (used for SVM training).}
\label{tab:networks}

\resizebox{0.46\textwidth}{!}{%

\begin{tabular}{cccccc}

& Conv & Model & Para- & Vector \\
Network & Layers & Size & meters & Size  \\ \hline

SqueezeNet (SQ) & 18 & 4.41MB & 1.24M & 1000  \\ \hline

MobileNetv2 (MB2) & 53 & 13MB & 3.5M  & 1280  \\ \hline

ResNet50 (R50) & 50 & 146MB & 25.6M  & 2048  \\ \hline

\multicolumn{5}{c}{} \\

\end{tabular}%
}

\end{table}

\begin{table}[]
\centering

\caption{Distribution of face images in the database across the different classes. F=Female. M=Male.}
\label{tab:db-breakdown}

\setlength{\tabcolsep}{2pt}

\resizebox{0.46\textwidth}{!}{%

\begin{tabular}{|c||cc||cc||cc||cc||cc|}


\cline{2-9}

\multicolumn{1}{c||}{} & \multicolumn{2}{c||}{\textbf{0-2 years old}} & \multicolumn{2}{c||}{\textbf{4-6 years old}} & \multicolumn{2}{c||}{\textbf{8-13 years old}} & \multicolumn{2}{c||}{\textbf{15-20 years old}} & \multicolumn{2}{c}{} \\ \cline{2-9}

\multicolumn{1}{c||}{fold} & \textbf{F / M} & \textbf{total} & \textbf{F / M} & \textbf{total} & \textbf{F / M} & \textbf{total} & \textbf{F / M} & \textbf{total} & \multicolumn{1}{c}{} &  \multicolumn{1}{c}{} \\ \hhline{=========} 

\textbf{1} & 202 / 188 & 390 & 226 / 130 & 356 & 122 / 40 & 162 & 82 / 28 & 110 &  \multicolumn{1}{c}{} & \multicolumn{1}{c}{} \\ \cline{1-9} 

\textbf{2} & 0 / 1 & 1 & 252 / 82 & 334 & 339 / 195 & 534 & 114 / 251 & 365 &  \multicolumn{1}{c}{}  &  \multicolumn{1}{c}{} \\ \cline{1-9}

\textbf{3} & 164 / 277 & 441 & 93 / 165 & 258 & 288 / 57 & 345 & 114 / 46 & 160 &  \multicolumn{1}{c}{} &  \multicolumn{1}{c}{} \\ \cline{1-9}

\textbf{4} & 3 / 48 & 51 & 163 / 33 & 196 & 98 / 307 & 405 & 197 / 117 & 314 &  \multicolumn{1}{c}{}  &  \multicolumn{1}{c}{} \\ \cline{1-9}

\textbf{5} & 115 / 5 & 120 & 151 / 251 & 402 & 98 / 121 & 219 & 101 / 45 & 146 &  \multicolumn{1}{c}{}  &  \multicolumn{1}{c}{} \\ \hhline{=========} 

\textbf{all} & \textbf{484} / \textbf{519} & \textbf{1003} & \textbf{885} / \textbf{661} & \textbf{1546} & \textbf{945} / \textbf{720} & \textbf{1665} & \textbf{608} / \textbf{487} & \textbf{1095} &  \multicolumn{1}{c}{} &  \multicolumn{1}{c}{} \\ \cline{1-9}

\multicolumn{11}{c}{}  \\


\cline{2-11}
 
\multicolumn{1}{c||}{} & \multicolumn{2}{c||}{\textbf{25-32 years old}} & \multicolumn{2}{c||}{\textbf{38-43 years old}} & \multicolumn{2}{c||}{\textbf{48-53 years old}} & \multicolumn{2}{c||}{\textbf{60-99 years old}} & \multicolumn{2}{c|}{\textbf{ALL age groups}} \\ \cline{2-11}

\multicolumn{1}{c||}{fold} & \textbf{F / M} & \textbf{total} & \textbf{F / M} & \textbf{total} & \textbf{F / M} & \textbf{total} & \textbf{F / M} & \textbf{total} & \textbf{F / M} & \textbf{total} \\ \hhline{===========} 

\textbf{1} & 478 / 646 & 1124 & 162 / 239 & 401 & 58 / 103 & 161 & 47 / 41 & 88 & 1377 / 1415 & 2792 \\ \hline

\textbf{2} & 204 / 228 & 432 & 129 / 179 & 308 & 37 / 50 & 87 & 49 / 45 & 94 & 1124 / 1031 & 2155 \\ \hline

\textbf{3} & 317 / 195 & 512 & 89 / 117 & 206 & 54 / 22 & 76 & 52 / 73 & 125 & 1171 / 952 & 2123 \\ \hline

\textbf{4} & 429 / 201 & 630 & 145 / 191 & 336 & 38 / 32 & 70 & 54 / 25 & 79 & 1127 / 954 & 2081 \\ \hline

\textbf{5} & 314 / 286 & 600 & 188 / 139 & 327 & 90 / 67 & 157 & 90 / 87 & 177 & 1147 / 1001 & 2148 \\ \hhline{===========} 

\textbf{all} & \textbf{1742} / \textbf{1556} & \textbf{3298} & \textbf{713} / \textbf{865} & \textbf{1578} & \textbf{277} / \textbf{274} & \textbf{551} & \textbf{292} / \textbf{271} & \textbf{563} & \textbf{5946} / \textbf{5353} & \textbf{11299} \\ \hline

\multicolumn{11}{c}{} \\

\end{tabular}%
}

\end{table}

\section{EXPERIMENTAL FRAMEWORK}
\label{sec:exp_framework}

\subsection{Database}
\label{sec:db}

We use the Adience benchmark \cite{[EidingerHassner14_Adience]}, designed for age and gender classification.
It comprises unconstrained smartphone images from Flickr, exhibiting substantial variability in pose and lighting (Figure~\ref{fig:adience_images}).
From the original 26,580 images of 2,284 subjects, we use the frontal subset of \cite{[Alonso21ietSoftBio]} which removed non-frontal images to simulate selfie captures, comprising 11,299 images (5,946 females and 5,353 males). The class distribution is given in Table~\ref{tab:db-breakdown}.
The age distribution is imbalanced across groups and folds, which may affect classification performance and variability.
Facial landmarks are detected using MTCNN \cite{[Zhang16]}. Images are rotated to align the eye axis and scaled to the average inter-eye distance of 105 pixels.
A 224$\times$224 face crop is extracted around the landmark centroid, together with two 105$\times$105 ocular crops centred on each eye. All crops are resized to 113$\times$113 for CNN input.
Left eye crops are horizontally flipped to enforce a common canonical orientation, consistent with the recognition networks used in our experiments \cite{2025_AlonsoIWAIPR_Ocular_Crops_Biometrics}.


\begin{table}[t]
\centering

\caption{\textbf{Gender estimation}: Overall accuracy using pre-trained CNN models and SVM classification with different CNN initialisations and SVM normalisation methods. Rows \textit{gain} refer to the accuracy increase when initialising the CNN in the same domain vs in a different domain.}
\label{tab:gender-accuracy-svm}

\setlength{\tabcolsep}{2pt}

\begin{adjustbox}{width=0.48\textwidth}

\begin{tabular}{|r|cc|cc|cc|}

\multicolumn{1}{c}{} & \multicolumn{6}{c}{\textbf{GENDER (Pre-trained CNN + SVM)}} \\ 

\cline{2-7}

\multicolumn{1}{c|}{} & \multicolumn{2}{c|}{\textbf{Full Face}} & \multicolumn{2}{c|}{\textbf{Ocular}} & \multicolumn{2}{c|}{\textbf{Ocular L+R}} \\  \hline
 
\textbf{Net + init} & Stand. & L2-norm & Stand. & L2-norm & Stand. & L2-norm \\  \hline \hline
 
\textbf{SQ} face & 82.5 ± 0.8 & \textbf{82.6} ± 1.9 & 66.7 ± 1.4 & 65 ± 1.1 & 69.3 ± 1.3 & 67.3 ± 2 \\ 

 ocular & 72.6 ± 2.2 & 70.7 ± 2 & 77.3 ± 1.1 & \textbf{77.9} ± 2.1 & 79.4 ± 1.4 & \textbf{79.8} ± 2.5 \\ \cline{2-7}

 \textit{gain} & 9.9 & 11.9 & 10.6 & 12.9 & 10.1 & 12.5	
 \\ \hline \hline

\textbf{MB2} face & 76.9 ± 2.9 & \textbf{81.7} ± 2.6 & 65.5 ± 2.1 & 66.8 ± 2.1 & 68.6 ± 2.4 & 69.7 ± 2.2 \\ 

ocular & 74.2 ± 1.8 & 74.9 ± 1.8 & 75.7 ± 2 & \textbf{78.3} ± 1.7 & 77.8 ± 2.8 & \textbf{80} ± 1.7 \\ \cline{2-7}

 \textit{gain} & 2.7 & 6.8 & 10.2 & 11.5 & 9.2 & 10.3	
 \\ \hline \hline

\textbf{R50} face & 81.9 ± 2 & \textbf{84.8} ± 1.7 & 65.1 ± 1.9 & 65.8 ± 1.6 & 68 ± 1.9 & 68.7 ± 1.6 \\ 

ocular & 75.6 ± 2.2 & 75.5 ± 2.3 & 77.4 ± 1.5 & \textbf{80.7} ± 1.5 & 80.1 ± 1 & \textbf{82.9} ± 1.3 \\ \cline{2-7}

 \textit{gain} & 6.3 & 9.3 & 12.3 & 14.9 & 12.1 & 14.2	
 \\ \hline

 \multicolumn{7}{c}{} \\

\end{tabular}

\end{adjustbox}

\end{table}


\begin{table}[t]
\centering

\caption{\textbf{Age estimation}: Overall accuracy using pre-trained CNN models and SVM classification with different CNN initialisations and SVM normalisation methods. Rows \textit{gain} refer to the accuracy increase when initialising the CNN in the same domain vs in a different domain.}
\label{tab:age-accuracy-svm}

\setlength{\tabcolsep}{2pt}

\begin{adjustbox}{width=0.48\textwidth}

\begin{tabular}{|c|cc|cc|cc|}

\multicolumn{1}{c}{} & \multicolumn{6}{c}{\textbf{AGE: EXACT GROUP ACCURACY (Pre-trained CNN + SVM)}} \\ 

\cline{2-7}

\multicolumn{1}{c|}{} & \multicolumn{2}{c|}{\textbf{Full Face}} & \multicolumn{2}{c|}{\textbf{Ocular}} & \multicolumn{2}{c|}{\textbf{Ocular L+R}} \\  \hline
 
\textbf{Net + init} & Stand. & L2-norm & Stand. & L2-norm & Stand. & L2-norm \\  \hline \hline

\textbf{SQ} face & 47.2 ± 6.1 & \textbf{49.1} ± 5.8 & 35.1 ± 3.1 & 35.2 ± 3.6 & 38.5 ± 3.9 & 38.2 ± 3.8 \\ 

 ocular & 41 ± 4 & 40.9 ± 4.5 & 49.5 ± 5.3 & \textbf{51.9} ± 5 & 52.8 ± 6.2 & \textbf{54.6} ± 5.7 \\ \cline{2-7}

 \textit{gain} & 6.2 & 8.2 & 14.4 & 16.7 & 14.3 & 16.4	
 \\ \hline \hline

\textbf{MB2} face & 49.2 ± 5 & \textbf{54.1} ± 5.5 & 33.3 ± 3.4 & 34.2 ± 4.5 & 36.8 ± 3.5 & 36.6 ± 4.9 \\ 

 ocular & 39 ± 4.3 & 42 ± 4.6 & 46.7 ± 2.6 & \textbf{53.4} ± 4.4 & 50.1 ± 3 & \textbf{56.6} ± 4.6 \\ \cline{2-7}

 \textit{gain} & 10.2 & 12.1 & 13.4 & 19.2 & 13.3 & 20	
 \\ \hline \hline

\textbf{R50} face & 52.5 ± 3.9 & \textbf{57.6} ± 4.1 & 33.1 ± 3.7 & 35.3 ± 4.9 & 35.8 ± 4.9 & 38.4 ± 5.4 \\ 

 ocular & 40.4 ± 4.2 & 41 ± 5 & 48.6 ± 3.2 & \textbf{53.1} ± 4.5 & 51.9 ± 3.6 & \textbf{56.6} ± 4.3 \\ \cline{2-7}

 \textit{gain} & 12.1 & 16.6 & 15.5 & 17.8 & 16.1 & 18.2	
 \\ \hline 

\multicolumn{7}{c}{} \\

\end{tabular}

\end{adjustbox}

\begin{adjustbox}{width=0.48\textwidth}

\begin{tabular}{|r|cc|cc|cc|}

\multicolumn{1}{c}{} & \multicolumn{6}{c}{\textbf{AGE: 1-OFF GROUP ACCURACY (Pre-trained CNN + SVM)}} \\ 

\cline{2-7}

\multicolumn{1}{c|}{} & \multicolumn{2}{c|}{\textbf{Full Face}} & \multicolumn{2}{c|}{\textbf{Ocular}} & \multicolumn{2}{c|}{\textbf{Ocular L+R}} \\  \hline
 
\textbf{Net + init} & Stand. & L2-norm & Stand. & L2-norm & Stand. & L2-norm \\  \hline \hline

\textbf{SQ} face & 86.8 ± 3.1 & \textbf{87.7} ± 3.2 & 69.4 ± 3 & 67.8 ± 2.5 & 72.7 ± 3.7 & 71.1 ± 3 \\ 

 ocular & 77.3 ± 3 & 75.7 ± 3.4 & 84.2 ± 2.6 & \textbf{86} ± 2.7 & 87.1 ± 3.1 & \textbf{88} ± 3.4 \\ \cline{2-7}

 \textit{gain} & 9.5 & 12 & 14.8 & 18.2 & 14.4 & 16.9	
 \\ \hline \hline

\textbf{MB2} face & 89 ± 2.1 & \textbf{92.1} ± 1.9 & 68.3 ± 1.9 & 68.2 ± 2.5 & 71.8 ± 2.5 & 71 ± 2.6 \\ 

 ocular & 76.8 ± 3.9 & 77.7 ± 1.8 & 86.6 ± 2.1 & \textbf{89.8} ± 1.4 & 89.1 ± 2.2 & \textbf{91.5} ± 1.5 \\ \cline{2-7}

 \textit{gain} & 12.2 & 14.4 & 18.3 & 21.6 & 17.3 & 20.5	
 \\ \hline \hline

\textbf{R50} face & 91.2 ± 2.1 & \textbf{93.7} ± 1.5 & 68.7 ± 3 & 69.4 ± 2.6 & 72.3 ± 3.6 & 73.2 ± 3 \\ 

 ocular & 78 ± 3.8 & 77.6 ± 2.9 & 87.9 ± 2.2 & \textbf{89.8} ± 1.4 & 90.2 ± 1.9 & \textbf{92} ± 1.4 \\ \cline{2-7}

 \textit{gain} & 13.2 & 16.1 & 19.2 & 20.4 & 17.9 & 18.8	
 \\ \hline 

 \multicolumn{7}{c}{} \\

\end{tabular}

\end{adjustbox}

\end{table}

\subsection{Protocol}
\label{sec:protocol}

The Adience benchmark specifies a predefined five-fold protocol with disjoint Flickr albums across training and test sets.
Although each album may contain multiple identities, the ground truth provides subject IDs for each image, with no identities shared across albums. Thus, the folds are also subject-disjoint.
%
%
%
For each test fold, the remaining four folds are used for training and validation. Results are reported as mean accuracy and standard deviation across the five folds.
Following \cite{[EidingerHassner14_Adience]}, we also report 1-off age accuracy, which considers predictions within one age group of the true class as correct.
Face training images are augmented with horizontal mirroring and gamma correction ($\gamma$=0.5, 1, 1.5), following \cite{[Alonso21ietSoftBio]}, yielding six samples per original face image to mitigate overfitting and illumination changes.
For ocular images, the eye crops undergo the same gamma transformations, also giving six training eye samples per face image. 
Thus, both modalities have the same number of training samples.

Soft-biometrics classification is done using two approaches (Figure~\ref{fig:system}): $i$) a linear SVM \cite{[Vapnik95]} with CNN-extracted features, and $ii$) CNNs trained end-to-end. 
SVMs use identity-pretrained CNN features from the layer preceding the classifier (Table~\ref{tab:networks}), whereas end-to-end models additionally evaluate with scratch and ImageNet initialisations.
%
%
%
SVMs are trained with MATLAB's \texttt{fitclinear} with default settings, except for \texttt{PostFitBias=true}. 
For age estimation, a one-vs-one multi-class strategy with \(N(N-1)/2\) binary SVMs is used, with the class determined by majority voting.
%
%
In end-to-end training, batch normalisation and 50\% dropout are applied before the fully connected layer.
The networks are trained with cross-entropy loss and SGDM with momentum 0.9, batch size 128, learning rate \(10^{-3}\) multiplied by 0.1 every two epochs, L2 regularisation \(10^{-4}\), and up to 15 epochs with early stopping based on validation loss (patience 5).
The model with the lowest validation loss is retained. 
Experiments are done in MATLAB R2024b on an NVIDIA RTX 2080 Ti GPU.

We use the same Adience subset, augmentation, and five-fold protocol as in our previous study \cite{[Alonso21ietSoftBio]}, but adopt a stricter validation.
For each fold, training subjects are split 90\%/10\% into subject-disjoint training and validation subsets, stratified by age and gender, with selection also ensuring an approximate 90\%/10\% image ratio within each age-gender group. 
In contrast, \cite{[Alonso21ietSoftBio]} reserved 20\% of images per class for validation without explicitly enforcing subject-disjoint subsets.

\section{RESULTS}
\label{sec:results}

\subsection{Models pre-trained for identity}
\label{sec:results-pretrained}

The CNNs are pre-trained for face or ocular identity recognition on two large databases.
All identity-pretrained CNNs were initially trained on ImageNet.
%
%
%
For face recognition \cite{alonso25PRLdeepnetworkpruning}, 
the models were subsequently trained on 
the RetinaFace-cleaned MS1M dataset \cite{[Guo16_MSCeleb1M]} (5.1M images, 93.4K identities) and then fine-tuned on VGGFace2 \cite{[Cao18vggface2]} (3.14M images, 8631 identities). 
For ocular recognition, 
the models were trained on 1.9M ocular crops from VGGFace2, following \cite{2025_AlonsoIWAIPR_Ocular_Crops_Biometrics}.
%
%

Tables~\ref{tab:gender-accuracy-svm}, \ref{tab:age-accuracy-svm} show gender and age accuracies using fixed CNN features and SVM classification.
We compare feature-wise standardisation to zero mean and unit variance typical of SVMs (fitted on training folds), and sample-wise L2 normalisation (which does not require training).
%
%
Results are reported using either the whole face or the ocular region as input. 
`Ocular' columns refer to the left and right eye classified individually, whereas `Ocular L+R' averages both eye descriptors before SVM classification.
%


\begin{table}[t]
\centering

\caption{\textbf{Gender estimation}: Overall accuracy using CNN models trained for gender classification with different CNN initialisations and with all layers either unfrozen for end-to-end fine-tuning, or frozen except the final fully connected layer. Rows \textit{gain} refer to the accuracy increase when initialising the CNN in the same domain vs a different domain.}
\label{tab:gender-accuracy-cnn}

\setlength{\tabcolsep}{2pt}

\begin{adjustbox}{width=0.48\textwidth}

\begin{tabular}{|r|cc|cc|cc|}

\multicolumn{1}{c}{} & \multicolumn{6}{c}{\textbf{GENDER (CNN end-to-end)}} \\ 

\cline{2-7}

\multicolumn{1}{c|}{} & \multicolumn{2}{c|}{\textbf{Full Face}} & \multicolumn{2}{c|}{\textbf{Ocular}} & \multicolumn{2}{c|}{\textbf{Ocular L+R}} \\  \hline
 
\textbf{Net + init} & Frozen & Unfrozen & Frozen & Unfrozen & Frozen & Unfrozen \\  \hline \hline

\textbf{SQ} scratch 
 & - & 65.4 ± 2.3
 & - & 59.2 ± 2.5
 & - & 60.9 ± 2.5
 \\ \cline{2-7}

 ImageNet 
 & - & 67.1 ± 2.3
 & - & 59.9 ± 2.3
 & - & 62.2 ± 2.5
 \\ \hline
 
 face & 82.4 ± 1.9 & \textbf{85.0} ± 2.4 & 63.1 ± 2.0 & 75.3 ± 1.5 & 65.7 ± 1.6 & 77.3 ± 1.5 \\ \cline{2-7}

 ocular & 69.5 ± 1.7 & 80.6 ± 2.0 & 77.9 ± 1.9 & \textbf{78.9} ± 2.2 & 79.8 ± 2.1 & \textbf{80.5} ± 1.8 \\ \cline{2-7}

 \textit{gain} &  12.9 & 4.4 & 14.8 & 3.6 & 14.1 & 3.2	
 \\ \hline \hline

\textbf{MB2} scratch 
 & - & 62.8 ± 1.8
 & - & 56.3 ± 2.0
 & - & 58.1 ± 1.8
 \\ \cline{2-7}

 ImageNet 
 & - & 80.2 ± 1.8
 & - & 73.7 ± 1.7
 & - & 76.2 ± 1.4
 \\ \hline

 face & 76.7 ± 1.6 & \textbf{87.5} ± 1.2 & 63.3 ± 2.3 & 75.5 ± 1.8 & 65.8 ± 2.4 & 77.5 ± 1.7 \\ \cline{2-7}

 ocular & 73.6 ± 1.9 & 82.9 ± 2.2 & 77.7 ± 2.9 & \textbf{81.2} ± 2.8 & 79.6 ± 2.7 & \textbf{82.5} ± 3.0  \\ \cline{2-7}

 \textit{gain} &  3.1 & 4.6 & 14.4 & 5.7 & 13.8 & 5	
 \\ \hline \hline

 \textbf{R50} scratch 
 & - & 67.6 ± 3.2
 & - & 58.7 ± 2
 & - & 60.1 ± 2.4
 \\ \cline{2-7}

 ImageNet 
 & - & 81.9 ± 3.2
 & - & 73.8 ± 2
 & - & 76 ± 2.1
 \\ \hline

 face & 82.6 ± 1.2 & \textbf{88.1} ± 1.7 & 61.9 ± 1.9 & 76.1 ± 2.3 & 64.0 ± 2.3 & 78.5 ± 2.2 \\ \cline{2-7}

 ocular & 72.7 ± 3.4 & 82.9 ± 2.5 & 80.2 ± 1.8 & \textbf{81.3} ± 1.7 & 81.7 ± 2.1 & \textbf{82.8} ± 1.7 \\ \cline{2-7}

 \textit{gain} &  9.9 & 5.2 & 18.3 & 5.2 & 17.7 & 4.3	 
 \\ \hline


\end{tabular}

\end{adjustbox}

\end{table}


\begin{table}[t]
\centering

\caption{\textbf{Age estimation}: Overall accuracy using CNN models trained for age classification with different CNN initialisations and with all layers either unfrozen for end-to-end fine-tuning, or frozen except the final fully connected layer. Rows \textit{gain} refer to the accuracy increase when initialising the CNN in the same domain vs a different domain.}
\label{tab:age-accuracy-cnn}

\setlength{\tabcolsep}{2pt}

\begin{adjustbox}{width=0.48\textwidth}

\begin{tabular}{|r|cc|cc|cc|}

\multicolumn{1}{c}{} & \multicolumn{6}{c}{\textbf{AGE: EXACT GROUP ACCURACY (CNN end-to-end)}} \\ 

\cline{2-7}

\multicolumn{1}{c|}{} & \multicolumn{2}{c|}{\textbf{Full Face}} & \multicolumn{2}{c|}{\textbf{Ocular}} & \multicolumn{2}{c|}{\textbf{Ocular L+R}} \\  \hline
 
\textbf{Net + init} & Frozen & Unfrozen & Frozen & Unfrozen & Frozen & Unfrozen \\  \hline \hline

\textbf{SQ} scratch 
 & - & 35.3 ± 6.7
 & - & 34.2 ± 5.4
 & - & 35.6 ± 5.8
 \\ \cline{2-7}

 ImageNet 
 & - & 37.0 ± 8.1
 & - & 35.7 ± 7.2
 & - & 36.7 ± 7.4
 \\ \hline
 
 face & 48.6 ± 4.5 & \textbf{52.7} ± 3.1 & 33.1 ± 4.9 & 44.5 ± 4.6 & 34.7 ± 5.6 & 47.0 ± 5.2 \\  \cline{2-7}

 ocular & 40.1 ± 5.0 & 48.8 ± 4.8 & 51.6 ± 4.9 & \textbf{52.6} ± 5.6 & 54.0 ± 5.3 & \textbf{55.4} ± 6.0  \\ \cline{2-7}

 \textit{gain} &  8.5 & 3.9 & 18.5 & 8.1 & 19.3 & 8.4	
 \\ \hline \hline

 \textbf{MB2} scratch 
 & - & 33.2 ± 6.3
 & - & 33.6 ± 6.4
 & - & 34.5 ± 6.6
 \\ \cline{2-7}

 ImageNet 
 & - & 48.5 ± 5.4
 & - & 44.7 ± 4.5
 & - & 47.1 ± 4.9
 \\ \hline

 face & 51.8 ± 4.3 & \textbf{56.5} ± 4.8 & 32.8 ± 4.7 & 46.6 ± 5.4 & 34.4 ± 6.2 & 49.5 ± 6.0 \\ \cline{2-7}

 ocular & 40.7 ± 6.4 & 51.2 ± 4.6 & 52.7 ± 3.9 & \textbf{55.7} ± 4.0 & 55.2 ± 3.6 & \textbf{57.7} ± 4.1  \\ \cline{2-7}

 \textit{gain} &  11.1 & 5.3 & 19.9 & 9.1 & 20.8 & 8.2	
 \\ \hline \hline

\textbf{R50} scratch 
 & - & 34.2 ± 3.8
 & - & 36.0 ± 4.0
 & - & 36.7 ± 4.5
 \\ \cline{2-7}

 ImageNet 
 & - & 48.7 ± 5.8
 & - & 44 ± 6.5
 & - & 46.5 ± 7.1
 \\ \hline

 face & 55.5 ± 5.2 & \textbf{56.8} ± 5.3 & 33.7 ± 5.2 & 45.6 ± 4.1 & 36.7 ± 6.4 & 48.3 ± 4.7 \\  \cline{2-7}

 ocular & 40.0 ± 4.9 & 50.2 ± 4.7 & 52.9 ± 4.7 & \textbf{53.8} ± 5.1 & 55.9 ± 4.9 & \textbf{56.4} ± 5.7  \\ \cline{2-7}

 \textit{gain} &  15.5 & 6.6 & 19.2 & 8.2 & 19.2 & 8.1	
 \\ \hline

\multicolumn{7}{c}{} \\

\end{tabular}

\end{adjustbox}

\begin{adjustbox}{width=0.48\textwidth}

\begin{tabular}{|r|cc|cc|cc|}

\multicolumn{1}{c}{} & \multicolumn{6}{c}{\textbf{AGE: 1-OFF GROUP ACCURACY (CNN end-to-end)}} \\ 

\cline{2-7}

\multicolumn{1}{c|}{} & \multicolumn{2}{c|}{\textbf{Full Face}} & \multicolumn{2}{c|}{\textbf{Ocular}} & \multicolumn{2}{c|}{\textbf{Ocular L+R}} \\  \hline
 
\textbf{Net + init} & Frozen & Unfrozen & Frozen & Unfrozen & Frozen & Unfrozen \\  \hline \hline

\textbf{SQ} scratch 
 & - & 69.7 ± 4.1
 & - & 68 ± 3.0
 & - & 70.0 ± 3.8
 \\ \cline{2-7}

 ImageNet 
 & - & 70.4 ± 4.2
 & - & 69.2 ± 4.5
 & - & 70.9 ± 5.3
 \\ \hline
 
 face & 85.2 ± 4.0 & \textbf{90.4} ± 2.7 & 63.8 ± 3.2 & 79.6 ± 2.9 & 66.1 ± 4.0 & 81.8 ± 3.1 \\  \cline{2-7}

 ocular & 73.3 ± 3.6 & 84.8 ± 3.7 & 84.1 ± 3.6 & \textbf{86.7} ± 4.1 & 85.7 ± 3.8 & \textbf{88.4} ± 4.2  \\ \cline{2-7}

 \textit{gain} & 11.9 & 5.6 & 20.3 & 7.1 & 19.6 & 6.6	
 \\ \hline \hline

 \textbf{MB2} scratch 
 & - & 66.3 ± 4.1
 & - & 66.9 ± 3.8
 & - & 68.8 ± 4.2
 \\ \cline{2-7}

 ImageNet 
 & - & 85.9 ± 3.6
 & - & 79.8 ± 3.6
 & - & 81.4 ± 3.8
 \\ \hline

 face & 88.6 ± 2.9 & \textbf{93.5} ± 2.7 & 64.6 ± 3.2 & 81.8 ± 4.6 & 67.3 ± 4.7 & 84.1 ± 5.0 \\ \cline{2-7}

 ocular & 74.2 ± 3.3 & 87.9 ± 3.1 & 87.7 ± 2.2 & \textbf{90.6} ± 2.3 & 89.5 ± 2.1 & \textbf{91.9} ± 2.0  \\ \cline{2-7}

 \textit{gain} & 14.4 & 5.6 & 23.1 & 8.8 & 22.2 & 7.8	
 \\ \hline \hline

 \textbf{R50} scratch 
 & - & 67.3 ± 2.4
 & - & 69.4 ± 2.1
 & - & 70.6 ± 2.9
 \\ \cline{2-7}

 ImageNet 
 & - & 85.3 ± 3.2
 & - & 79.5 ± 4.8
 & - & 81.4 ± 4.8
 \\ \hline

 face & 91.1 ± 2.1 & \textbf{94.0} ± 1.1 & 66.1 ± 3.6 & 83.2 ± 3.4 & 69.8 ± 4.5 & 85.1 ± 3.6 \\ \cline{2-7}

 ocular & 74.0 ± 3.3 & 86.7 ± 2.8 & 88.3 ± 1.8 & \textbf{90.5} ± 2.9 & 90.1 ± 1.7 & \textbf{91.9} ± 3.0  \\ \cline{2-7}

 \textit{gain} & 17.1 & 7.3 & 22.2 & 7.3 & 20.3 & 6.8	
 \\ \hline

\end{tabular}

\end{adjustbox}

\end{table}

The results show a clear dependency between the pre-training domain and the input region.
Networks pre-trained for face recognition perform better with full-face images, whereas ocular-recognition pre-training performs better with ocular images.
This is particularly pronounced for ocular data, especially in age estimation (Table \ref{tab:age-accuracy-svm}). 
For example, with MB2/R50 and L2-normalised features, ocular pre-training improves exact and 1-off ocular age accuracy by 18-22\%.
In ocular gender estimation (Table \ref{tab:gender-accuracy-svm}), improvements of 10-15\% are also observed with MB2/R50 and same-domain pre-training. 

Combining both eyes also consistently improves performance for both gender and age. 
Regarding feature scaling, L2 normalisation generally outperforms standardisation and requires no additional training.
This is consistent with the representation employed by modern recognition networks, in which feature embeddings are L2-normalised, and discrimination is based primarily on angular relationships, such as cosine similarity \cite{2018_wangCVPR_CosFace_Margin_FR}.
%
%
It can also be observed that the performance gap between face and ocular L+R is small: 
84.8\% vs 82.9\% for gender, 57.6\% vs 56.6\% for exact age, and 93.7\% vs 92.0\% for 1-off age, respectively.
These results suggest that the ocular region retains much of the demographic information available in the full face, provided that pre-training matches the input region and both eyes are combined.

\begin{figure*} [!t]
\centering
\includegraphics[width=0.94\textwidth]{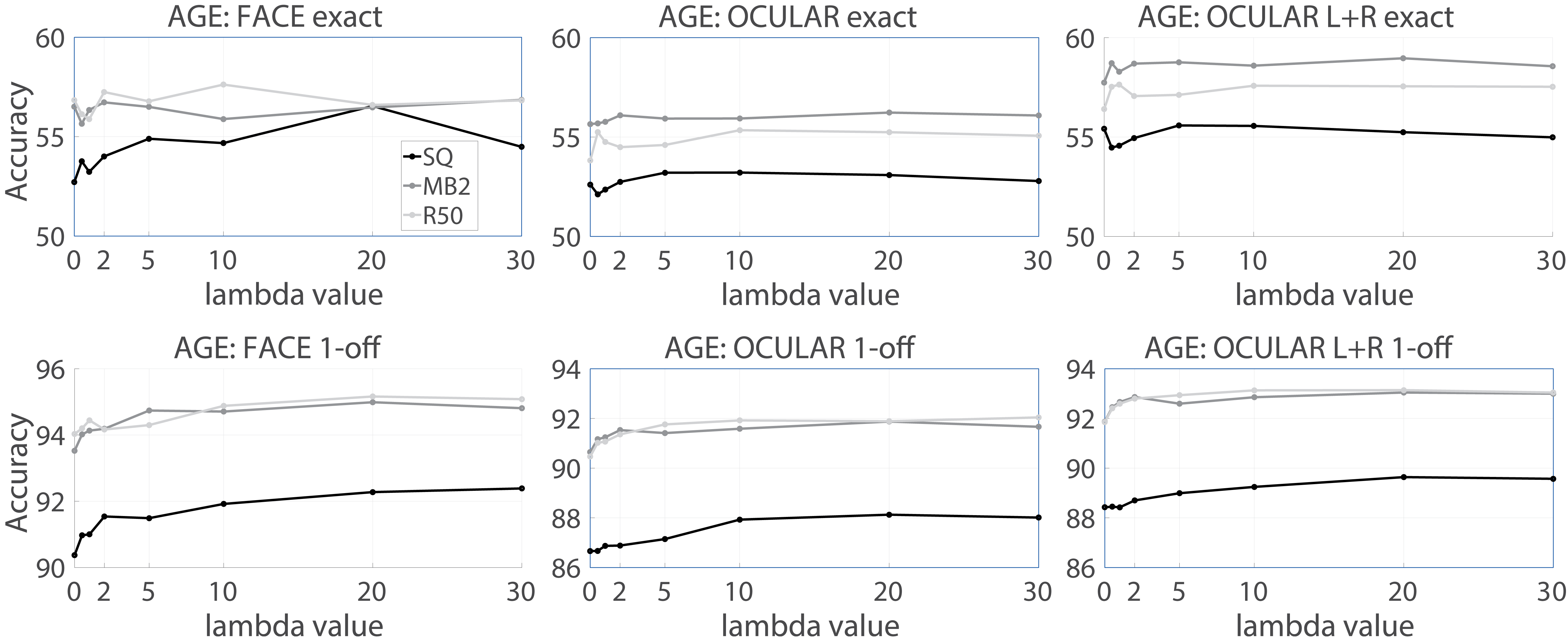}
\caption{Effect of $\lambda_{\mathrm{ord}}$ on age estimation accuracy, averaged over the five test folds. $\lambda_{\mathrm{ord}}=0$ corresponds to standard cross-entropy without the ordinal term.}
\label{fig:age_ordinal_loss}
\end{figure*}

\subsection{Fine-tuning for soft-biometrics}
\label{sec:results-fine-tune}

We compare two strategies (Tables~\ref{tab:gender-accuracy-cnn}, \ref{tab:age-accuracy-cnn}): 'Frozen', where only the last classification layer is trained, and 'Unfrozen', allowing end-to-end fine-tuning. 
The first is conceptually similar to the SVM experiments but with a neural network classifier.
%
%
Unlike the SVM experiments, 'Ocular L+R' averages the softmax scores of both eyes rather than their descriptors.
To assess the benefit of identity pre-training, we also evaluate direct ImageNet initialisation and training from scratch in the 'Unfrozen' setting.

Matching the pre-training and input domains is important, particularly with frozen features, yielding gains of up to 18.3\%, 20.8\%, and 23.1\% for gender, exact age, and 1-off age, respectively. 
These gains decrease to approximately 3-9\% with end-to-end fine-tuning, indicating reduced domain dependence if the network is allowed to adapt its representations. 
Nevertheless, the best accuracies are obtained with end-to-end training and matched pre-training domains.
In all unfrozen configurations, ImageNet initialisation outperforms scratch, while matched identity pre-training further improves accuracy over ImageNet, in some cases by 10-20\%, especially in age estimation.
Since the identity networks themselves start from ImageNet, this highlights the benefit of the additional face/ocular identity pre-training stage. 
Combining both eyes consistently improves ocular performance as well, narrowing the gap with the full face, especially for age: 57.7\% vs 56.8\% exact and 91.9\% vs 94.0\% 1-off accuracy (ocular L+R vs face). 
For gender, the gap is larger (82.8\% vs 88.1\%).

\subsection{Ordinal loss for multi-class classification}
\label{sec:results-ordinal-loss}

The standard cross-entropy loss $\mathcal{L}_{CE} = -\log p_y$ ignores the ordering of age groups.
Previous t-SNE visualisations suggested a tendency for nearby age groups to occupy neighbouring regions in the learned feature space \cite{[Alonso21ietSoftBio]}, further motivating the exploitation of their ordinal structure.
We thus introduce an ordinal penalty 
that increases with the distance from the true age group $y$ ($y=1,\ldots,8$):

\begin{equation}
\mathcal{L} = 
\mathcal{L}_{CE} +
\lambda_{\mathrm{ord}}
\sum_{j=1}^{K} p_j
\left(\frac{j-y}{K-1}\right)^2
\end{equation}

Here, $p_j$ is the predicted probability for age group $j$, and $K=8$ is the number of age groups. The squared distance penalises larger age-group errors, with maximum cost normalised to one.
The penalty operates on age-group indices rather than chronological age differences, since the groups have unequal age ranges.
Figure~\ref{fig:age_ordinal_loss} analyses the effect of $\lambda_{\mathrm{ord}}$. 
Increasing $\lambda_{\mathrm{ord}}$ generally improves 1-off test accuracy across networks and input regions, with less consistent effects on exact accuracy. 
This is expected, as the ordinal penalty prioritises reducing large age-group errors over exact classification.
Performance saturates around $\lambda_{\mathrm{ord}}=20$, which we select as a compromise between exact and 1-off metrics.
Table~\ref{tab:accuracy-compare-svm-cnn} compares these results with the best SVM and conventional end-to-end configurations of the previous two subsections.
With the selected $\lambda_{\mathrm{ord}}$, the ordinal loss yields higher 1-off age accuracy than conventional end-to-end training in all evaluated configurations, reaching 95.2\% for face and 93.1\% for ocular L+R, while exact accuracy improves in most configurations.


\begin{table*}[htb]
\centering

\caption{Summary of the best accuracies and comparison with previous work on Adience. Bold values indicate the highest accuracy achieved by our proposed methods for each network and task, while grey-shaded cells indicate the best accuracy reported in previous work. Ordinal loss results use $\lambda$=20. Comparisons are indicative because experimental protocols, age classes or number of images per fold may differ.}
\label{tab:accuracy-compare-svm-cnn}



\begin{adjustbox}{width=0.82\textwidth}

\begin{tabular}{|c|c||ccc|}

\multicolumn{2}{c}{} & \multicolumn{3}{c}{\textbf{FULL FACE}} \\ \hline

\textbf{Work} & \textbf{Method} & \multicolumn{1}{c}{\textbf{Gender}} & \multicolumn{1}{c}{\textbf{Age: exact}} & \multicolumn{1}{c||}{\textbf{Age: 1-off}} \\  \hline \hline
 
Proposed & SQ: SVM & 82.6 ± 1.9 &  49.1 ± 5.8 &  87.7 ± 3.2   \\ \cline{2-5}

 & SQ: end-to-end &  \textbf{85} ± 2.4 &  {52.7} ± 3.1 &  {90.4} ± 2.7 \\ \cline{2-5}

 & SQ: end-to-end (ordinal loss) &  -  & \textbf{56.5} ± 4.1
  &  \textbf{92.3} ± 1.7
 \\ \hline \hline

Proposed  & MB2: SVM & 81.7 ± 2.6  & 54.1 ± 5.5 & 92.1 ± 1.9   \\ \cline{2-5}

 & MB2: end-to-end &  \textbf{87.5} ± 1.2 &  \textbf{56.5} ± 4.8 &  {93.5} ± 2.7  \\ \cline{2-5}

 & MB2: end-to-end (ordinal loss) &  -  & \textbf{56.5} ± 5.3  &  \textbf{95} ± 2
 \\ \hline \hline

Proposed  & R50: SVM & 84.8 ± 1.7 & \textbf{57.6} ± 4.1 & 93.7 ± 1.5 \\ \cline{2-5}

 & R50: end-to-end &  \textbf{88.1} ± 1.7 &  56.8 ± 5.3 &  {94} ± 1.1  \\ \cline{2-5}

 & R50: end-to-end (ordinal loss) &  -  & 56.6 ± 5.1   &  \textbf{95.2} ± 0.9
 \\ \hline

\multicolumn{5}{c}{} \\ \hline 

\cite{[EidingerHassner14_Adience]} & LBP/FPLBP + SVM  & 77.8 ± 1.3 & 45.1 ± 2.6 & 79.5 ±1.4 \\ \hline

\cite{[LeviHassner15_AgeGender]} & Light CNN & 86.8 ± 1.4 & 50.7 ± 5.1 & 84.7 ± 2.2 \\ \hline

\cite{[Angeloni19iccvwAgeFaceParts]} & Multi-Stream CNN of face parts  & - & 51.03 ± 4.63  & 83.41 ± 3.17  \\ \hline  

\cite{[Bhattacharyya19scGenderFacialRegionsGA]} & compass LBP + SVM & 87.71 & - & - \\ \hline 

\cite{[Fang19neurocomputingGenderAgeMultiStage]} & VGG-19 with local/global relationship & 93.52 & - & -  \\ \hline

\cite{2019_HungICMR_MultipleFaceInformaticsModules} &  ProgressiveNet + PackNet  & 89.08 & 57.30  & -  \\ \hline 

\cite{[Gyawali20icccntFaceAgeMTCNN_VGG]} & VGG-Face  & - & 70.96 & 92.7  \\ \hline

\cite{2020_KhanSensors_Multitask_Face_Attributes_CNNs} &  RAG-MCFP-DCNNs & 93.6 & 69.4  & -  \\ \hline 

\cite{[Zhang20tcsvtFaceAgLSTM]} & ResNet + Attention LSTM  & - & 67.83 ± 2.98 & \cellcolor{gray!20} 97.53 ± 0.59 \\ \hline

\cite{[Alonso21ietSoftBio]} & MobileNetv2, ResNet50 & 85.3 ± 5.4 &   53.8 ± 4.8 &   93.4 ± 1.1 \\ \hline

\cite{2021_BenkaddourInformatica_CNNAgeGender} & Deep CNN   & 95.6 & \cellcolor{gray!20} 91.75  & -  \\ \hline 

\cite{2023_GhrbanJCSSP_gender_face_eyes} & Wavelet Scattering CNN  & 91.0 & -  & -   \\ \hline

\cite{2023_KuprashevichAIST_MiVOLOAgeGender} & ViT MiVOLO-D1 & 96.51 & 68.69 &  \\ \hline 

\cite{2024_SinghNCA_HybridTransformerSequencer} & ViT-B/32 + BiLSTM   & 96.56 $\pm$ 0.27 & 84.91 ± 0.84  & -  \\ \hline

\cite{2024_WarisMS_GenderClassification} & Deep CNN + Soft voting classifier   & \cellcolor{gray!20} 97.61 &  - & - \\ \hline

\end{tabular}

\end{adjustbox}

\begin{adjustbox}{width=0.9\textwidth}

\begin{tabular}{|c|c||cc||cc||cc|}

\multicolumn{1}{c}{} \\
\multicolumn{1}{c}{} \\

\multicolumn{2}{c}{} & \multicolumn{6}{c}{\textbf{OCULAR}} \\ \cline{3-8}

\multicolumn{2}{c||}{} & \multicolumn{2}{c||}{\textbf{Gender}} & \multicolumn{2}{c||}{\textbf{Age: exact}} & \multicolumn{2}{c|}{\textbf{Age: 1-off}} \\ 

\hline

\textbf{Work} & \textbf{Method} & \multicolumn{1}{c}{\textbf{Ocular}} & \multicolumn{1}{c||}{\textbf{Ocular L+R}} & \multicolumn{1}{c}{\textbf{Ocular}} & \multicolumn{1}{c||}{\textbf{Ocular L+R}}  & \multicolumn{1}{c}{\textbf{Ocular}} & \multicolumn{1}{c|}{\textbf{Ocular L+R}} \\  \hline \hline
 
Proposed: SQ & SVM & 77.9 ± 2.1 &  79.8 ± 2.5 & 51.9 ± 5 &  54.6 ± 5.7 & 86 ± 2.7 &   88 ± 3.4     \\ \cline{2-8}

 & end-to-end &  \textbf{78.9} ± 2.2 &  \textbf{80.5} ± 1.8 &    {52.6} ± 5.6 &  \textbf{55.4} ± 6 &  {86.7} ± 4.1 &  {88.4} ± 4.2 \\ \cline{2-8}

 & end-to-end (ordinal loss) &  - &  - &  \textbf{53.1} ± 4.6 & 55.2 ± 4.9  &  \textbf{88.1} ± 3.1 &  \textbf{89.6} ± 3.3
 \\ \hline \hline

Proposed: MB2  & SVM & 78.3 ± 1.7 &  80 ± 1.7 & 53.4 ± 4.4 &  56.6 ± 4.6 & 89.8 ± 1.4 &   91.5 ± 1.5  \\ \cline{2-8}

 & end-to-end &  \textbf{81.2} ± 2.8 &  \textbf{82.5} ± 3 &   {55.7} ± 4 &  {57.7} ± 4.1 &  {90.6} ± 2.3 &  {91.9} ± 2 \\ \cline{2-8}

 & end-to-end (ordinal loss) &  - &  - & \textbf{56.2} ± 3.9  & \textbf{59} ± 4.4  & \textbf{91.9} ± 1.7  &  \textbf{93} ± 1.6 \\ \hline \hline

Proposed: R50  & SVM & 80.7 ± 1.5 &  \textbf{82.9} ± 1.3  &  53.1 ± 4.5 &  {56.6} ± 4.3 & 89.8 ± 1.4 &  {92} ± 1.4 \\ \cline{2-8}

 & end-to-end &  \textbf{81.3} ± 1.7 &  82.8 ± 1.7 &  {53.8} ± 5.1 &  56.4 ± 5.7 &  {90.5} ± 2.9 &  91.9 ± 3 \\ \cline{2-8}

 & end-to-end (ordinal loss) &  - &  - &  \textbf{55.2} ± 4.8 &  \textbf{57.6} ± 4.9  & \textbf{91.9} ± 2.1  &  \textbf{93.1} ± 1.7 \\ \hline

\multicolumn{8}{c}{} \\ \hline

\cite{[Rattani17_AgeOcularIJCB]} & Light CNN & -  & -  & -  & 46.97 ± 2.9  &  - &  80.96 ± 1.09 \\ \hline  

\cite{[Bhattacharyya19scGenderFacialRegionsGA]} & compass LBP + SVM &  \cellcolor{gray!20} 84.06 &  83.27 & - & - & - & - \\ \hline 

\cite{[Alonso21ietSoftBio]} & MobileNetv2, ResNet50 & 76.6 $\pm$ 3.3 & 78.9 $\pm$ 3.7 &   \cellcolor{gray!20} 45.9 ± 4.3 &    \cellcolor{gray!20} 48.8 ± 5.1 &   \cellcolor{gray!20}83.1 ± 3.7 &  \cellcolor{gray!20} 86.2 ± 3.1 \\ \hline

\cite{2023_GhrbanJCSSP_gender_face_eyes} & Deep Wavelet CNN  & 72.13 & \cellcolor{gray!20} 83.34 & -  & -  & -  & -  \\ \hline

\end{tabular}

\end{adjustbox}

\end{table*}

\subsection{Comparison with previous works}
\label{sec:results-soa}

Table~\ref{tab:accuracy-compare-svm-cnn} also compares our results with previous work using the Adience database.
However, differences in image subsets, evaluation protocols, age-group definitions, and external training data prevent strict comparisons.
Thus, the results provide a contextual comparison rather than a strict ranking.

Most of the literature using Adience employs full-face images.
For gender estimation, our best face accuracy is 88.1\% with R50 and end-to-end training, 
improving the original 77.8\% baseline \cite{[EidingerHassner14_Adience]} and our previous results on the same image subset but with a different training regime \cite{[Alonso21ietSoftBio]}.
MB2 follows closely with 87.5\%, and the shallower SQ still achieves a competitive 85\%.
%
%
%
%
Values above 90\% have also been reported, although several of these works employ different experimental settings or additional architectural 
components.
For example, \cite{[Fang19neurocomputingGenderAgeMultiStage]} combines global and local information and their interaction through dedicated layers, using an 80/20 train-test partition instead of the predefined Adience folds.
The work \cite{2020_KhanSensors_Multitask_Face_Attributes_CNNs} uses probabilistic maps to classify image pixels into seven categories (hair, eyebrows, eyes, etc.), then extracts and concatenates CNN features from each category map, and performs 10-fold cross-validation.
In \cite{2021_BenkaddourInformatica_CNNAgeGender}, images from the same subjects appear in both training and test sets, potentially leading to optimistic accuracy estimates, 
while \cite{2023_GhrbanJCSSP_gender_face_eyes} evaluates a wavelet-scattering-based pipeline on a reduced subset of 8,888 Adience images. 
MiVOLO \cite{2023_KuprashevichAIST_MiVOLOAgeGender} uses Adience as a test set after training a transformer-based architecture of 27.4M parameters on 500,000 proprietary images from the external Lagenda database. 
%
%
In \cite{2024_SinghNCA_HybridTransformerSequencer}, they combine a pre-trained ViT-B/32 transformer with additional BiLSTM-based sequencer blocks. This is a substantially larger architecture, since a standard ViT-B/32 alone contains $\simeq$88M parameters, more than three times the count of a ResNet50.
Finally, \cite{2024_WarisMS_GenderClassification} combines learned CNN descriptors with a soft-voting ensemble of multiple machine-learning classifiers.

For face-based age, our best exact and 1-off accuracies are 57.6\% (R50 with SVM) and 95.2\% (R50 with ordinal loss), respectively. MB2 falls just behind, and the shallower SQ remains competitive.
The highest reported exact accuracies include \cite{2021_BenkaddourInformatica_CNNAgeGender} (91.75\%),
%
with subject overlap between training and test sets, and \cite{2024_SinghNCA_HybridTransformerSequencer} (84.91\%), with the very large ViT-B/32 + BiLSTM model. 
In addition, the work \cite{2021_BenkaddourInformatica_CNNAgeGender} employs 6 age groups instead of the original 8, merging 0-2 with 4-6 and 8-13 with 15-20.
The mentioned method of \cite{2023_KuprashevichAIST_MiVOLOAgeGender}, trained on 500,000 proprietary images, reports 68.69\% exact face age accuracy.
Other methods surpassing our approach include \cite{[Gyawali20icccntFaceAgeMTCNN_VGG]} (70.96\%), which uses an 80/20 train-test split for each age class rather than the predefined five-fold Adience protocol,
and \cite{[Zhang20tcsvtFaceAgLSTM]} (67.83\%), which combines a residual network with an attention-LSTM architecture, and importantly, age-specific pre-training on the large IMDB-WIKI-101 database.
In 1-off face age accuracy, our approach is only surpassed by \cite{[Zhang20tcsvtFaceAgLSTM]} among the compared works.

Previous work using only the ocular region of Adience is considerably more limited.
For gender estimation, we achieve 81.3\% on individual eyes, and 82.9\% when combining both eyes.
This improves our previous results on the same subset \cite{[Alonso21ietSoftBio]} by more than 4\%. 
Our ocular gender results also approach those of 
%
\cite{[Bhattacharyya19scGenderFacialRegionsGA]} (84.06\%, single eye) and 
\cite{2023_GhrbanJCSSP_gender_face_eyes} (83.34\%, both eyes), although these works use different image subsets and experimental protocols.
The largest improvement is in ocular age estimation, where, to the best of our knowledge, 
our best observed accuracies are the highest reported on Adience
%
%
56.2/59.0\% exact and 91.9/93.1\% 1-off (single eye/both eyes).
%
%
Compared with \cite{[Alonso21ietSoftBio]} on the same subset, exact accuracy improves by over 10\% and 1-off accuracy by 7-9\%, also surpassing earlier results of \cite{[Rattani17_AgeOcularIJCB]}.
%
%
%

\section{CONCLUSIONS}
\label{sec:conclusion}

We investigated the transfer of identity-recognition representations to face and ocular gender and age estimation.
Three architectures (SqueezeNet, MobileNetv2, and ResNet50) were evaluated as fixed feature extractors with SVM classification and through task-specific end-to-end adaptation.
Experiments on 11,299 Adience smartphone images showed
%
the importance of matching the pre-training domain to the target image region. 
Face-recognition pre-training performs better for full-face inputs, whereas ocular-recognition pre-training provides larger gains for ocular images.
End-to-end fine-tuning mitigates domain mismatch, but matched identity pre-training still performs best, outperforming direct ImageNet initialisation and training from scratch.
Since both identity-pretrained variants originate from ImageNet, our results demonstrate the benefit of additional identity training for transferring representations to soft-biometric prediction.
We also introduce an ordinal loss for age estimation penalising age-group errors according to their distance from the true class.
The proposed ordinal loss yields higher observed 
1-off accuracy across architectures and regions, with smaller and less uniform gains in exact accuracy, consistent with its objective of discouraging predictions far from the correct age group.

Several additional observations emerge from our experiments.
For SVM classification, sample-wise L2 normalisation consistently outperforms feature-wise standardisation typically employed with SVMs, while combining both eyes improves both gender and age estimation.
Our strongest results are obtained with ResNet50 in most settings, but the smaller
MobileNetv2 and SqueezeNet remain competitive.
This suggests that useful soft-biometric representations do not necessarily require large architectures.
We also observed that the ocular region retains much of the demographic information available in the full face, particularly when 
pre-training and target domains match.
For gender, our best accuracies are 88.1\% (face) and 82.9\% (ocular, both eyes combined), whereas the difference is smaller for age.
With the ordinal loss, ocular (both eyes) achieves 59.0/93.1\% (exact/1-off age accuracy), compared with 56.6/95.2\% for the full face.
For ocular age, this improves previous results on the same Adience subset
\cite{[Alonso21ietSoftBio]} by more than 10\% in exact accuracy and by 7-9\% in 1-off accuracy.

Despite these gains, some previous face-based approaches report considerably higher accuracy, although often under different experimental settings, architectures, or training regimes.
This leaves substantial room for improvement with the ocular modality, particularly for exact age classification.
We will thus investigate the adaptation of more advanced face-based methods to ocular estimation, such as
%
%
explicitly modelling local and global information via dedicated layers \cite{[Fang19neurocomputingGenderAgeMultiStage]}, transformer and attention mechanisms \cite{[Zhang20tcsvtFaceAgLSTM],2023_KuprashevichAIST_MiVOLOAgeGender,2024_SinghNCA_HybridTransformerSequencer},
or using substantially larger training sets with soft-biometrics labels \cite{[Zhang20tcsvtFaceAgLSTM],2023_KuprashevichAIST_MiVOLOAgeGender}.
%
%
Finally, more sophisticated ordinal-learning formulations will be investigated to improve exact classification without sacrificing 1-off performance.
Evaluation on additional datasets and cross-database settings will also help assess generalisation.

\section*{\uppercase{Acknowledgements}}

\small

This work was partly done while F. A.-F. was a visiting researcher at the University of the Balearic Islands.
F. A.-F., K. H.-D., and J. B. thank the
Swedish Research Council (VR) and the EU (HORIZON Europe project PopEye, Grant Agreement 101168317). 
Funded by the EU. Views and opinions expressed are, however, those of the author(s) only and do not necessarily reflect those of the European Union or the European Research Executive Agency. Neither the European Union nor the granting authority can be held responsible for them.
This work is part of the Project PID2022-136779OB-C32 (PLEISAR) funded by MICIU/ AEI /10.13039/501100011033/ and FEDER, EU.
The paper used AI tools for text correction purposes only.


\bibliographystyle{apalike}
{\small

}

\end{document}